\documentclass[letterpaper,10 pt,conference]{ieeeconf}
\IEEEoverridecommandlockouts                              

\usepackage{amsmath}
\usepackage{array}
\usepackage{pdfpages}
\usepackage{cite}
\usepackage{graphicx}
\usepackage{url}
\usepackage{tabularx}
\usepackage{multirow}
\usepackage{amsmath, amsfonts}
\usepackage{xcolor}
\usepackage{pifont}

\newcolumntype{Y}{>{\centering\arraybackslash}X}

\begin{document}


\title{\LARGE \bf End2Race: An End-to-End Learning Framework \\ for Multi-Vehicle Autonomous Racing}


\author{Zhijie Qiao\textsuperscript{1,†}, Haowei Li\textsuperscript{1,†}, Zhong Cao\textsuperscript{1}, Henry X. Liu\textsuperscript{1,2,*}
\thanks{This research was partially funded by the DARPA TIAMAT Challenge (HR0011-24-9-0429).}%
\thanks{$^{1}$Z.~Qiao, H.~Li, Z.~Cao, and H.~X.~Liu are with the Department of Civil and Environmental Engineering, University of Michigan, Ann Arbor, MI 48109, USA.} 
\thanks{$^{2}$H.~X.~Liu is also with the University of Michigan Transportation Research Institute, Ann Arbor, MI 48109, USA. }
\thanks{†These authors contributed equally to this work.}%
\thanks{*Corresponding author: Henry X. Liu (henryliu@umich.edu).}%
}




\maketitle

\begin{abstract}

Autonomous racing serves as a compelling testbed for advancing autonomous vehicle systems. The 1/10-scale F1Tenth platform is widely adopted for education and research, with annual competitions held worldwide. Yet, leading solutions in these tournaments remain dominated by rule-based approaches. While learning-based methods have been proposed, they focus primarily on single-vehicle settings, leaving challenging multi-vehicle interactions largely unexplored. To address this gap, we introduce \textbf{End2Race}, an end-to-end learning framework designed for head-to-head autonomous racing. Using this framework, we develop a computationally efficient policy network that achieves an inference latency below 1~ms on an F1Tenth onboard computer. Extensive evaluations show that the policy generalizes to both novel tracks and unseen opponent behaviors, demonstrating high racing speeds, robust safety, and adaptive overtaking maneuvers. These results substantially outperform both prior learning-based methods and rule-based baselines. Codebase is available at https://github.com/michigan-traffic-lab/End2Race.

\end{abstract}

\IEEEpeerreviewmaketitle

\section{Introduction}

Autonomous racing combines perception, planning, and control under high-speed, dynamic, and computationally constrained conditions.
Enhancing algorithmic capabilities in this domain is therefore of interest to the broader development of autonomous driving.
F1Tenth~\cite{o2020f1tenth}, a 1/10-scale vehicular platform, provides an accessible testbed for autonomous racing education and research.
Equipped with a 2D LiDAR or an RGB-D camera, the vehicle can operate at speeds up to 17.9~m/s.
Annual competitions held worldwide attract broad participation from the community.

\begin{figure}[!t]
    \centering
    \vspace{0.15cm}
    \includegraphics[width=0.95\linewidth]{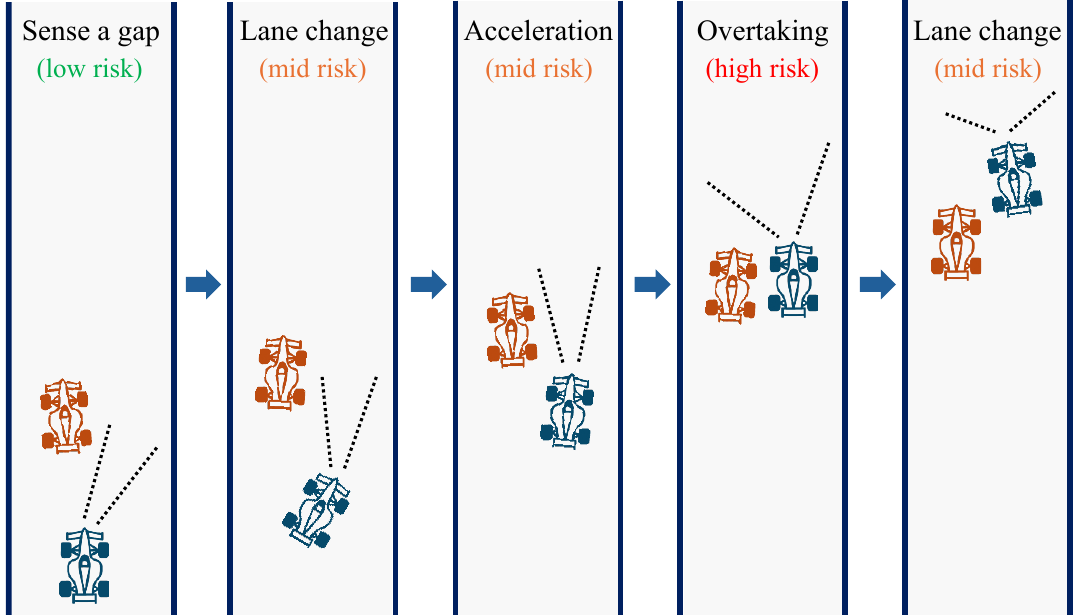}
    \caption{Illustration of the five-stage overtaking process.}
    \label{fig:example}
\end{figure}

Existing approaches for autonomous racing can be broadly grouped into map-based, reactive, and learning-based methods.
Among them, map-based pipelines are the most widely used, relying on a pre-constructed map to follow a globally optimized raceline.
Prior to racing, the vehicle is manually driven to map the track layout using SLAM~\cite{Macenski2021}.
A minimum-curvature raceline is then computed to serve as the global reference path~\cite{Heilmeier02102020}.
During racing, a particle filter~\cite{walsh2017cddt} tracks the vehicle's position, enabling a lattice planner~\cite{Pivtoraiko-2009-10176} to evaluate feasible local trajectories.
Motion controllers such as Pure Pursuit~\cite{coulter1992implementation} or model predictive control (MPC)~\cite{falcone2007predictive} then execute the selected trajectory.
However, map-based methods have two limitations.
First, they require preprocessing and cannot operate on novel tracks.
Second, the localization and planning pipeline incurs approximately 30~ms of computation on onboard hardware~\cite{zang2023localinn,muzzini2024gpu}.
Combined with the 25~ms sensor acquisition interval at 40~Hz, the resulting 55~ms total response delay causes the vehicle to travel up to half a meter before updated control commands take effect.
This leaves the vehicle susceptible to boundary collisions when transitioning from high-speed straights into sharp curves.

Reactive methods provide a faster, rule-based alternative. They operate directly on current sensor measurements and do not require map information, enabling operation on unknown tracks. The Follow-the-Gap algorithm~\cite{SEZER20121123}, for example, identifies the largest free region and steers toward its midpoint. However, these methods operate instantaneously on isolated observations with limited temporal context, leading to planning inconsistencies and control instability~\cite{hossain2022local}. As a result, they are rarely used in F1Tenth competitions.

Beyond rule-based methods, learning-based approaches have been widely studied. Early work focused on single-vehicle navigation~\cite{10186780,9811650}, whereas recent studies have shifted toward multi-vehicle interactions. Nevertheless, existing approaches face severe limitations, such as training on static obstacles and testing against dynamic opponents~\cite{zarrar2024tinylidarnet}, relying on a small set of human demonstrations~\cite{10275226}, or operating at impractically low racing speeds~\cite{steiner2025overtaking}. Importantly, none of these methods explicitly tackles high-speed, dynamic overtaking scenarios, which are intrinsically challenging and represent a key factor in the race outcome.
We illustrate a simplified five-stage overtaking process in Fig.~\ref{fig:example}.
First, the ego vehicle identifies an opening, executes a lane change to align with it, and accelerates rapidly to catch up with the opponent.
During the pass, the vehicles travel side by side for an extended period with minimal safety margins.
After completing the pass, the ego vehicle returns to the track center to regain lateral clearance.
Successful overtaking therefore depends on continuous decision-making, precise maneuver execution, rapid adaptation to the opponent, and minimal response time.
This requires sufficient model capacity and low inference latency, two conflicting demands that existing methods fail to reconcile.

To address these limitations, we propose \textbf{End2Race}, an end-to-end learning framework for real-time, head-to-head autonomous racing.
The framework supports the automatic generation of overtaking scenarios under diverse conditions, enabling systematic policy training and evaluation at scale.
Utilizing this framework, we train a racing policy based on a Gated Recurrent Unit (GRU) architecture~\cite{cho2014learning}, which takes a 360-degree 2D LiDAR scan and the current ego speed as inputs and directly outputs steering and speed commands.
On an F1Tenth onboard computer (NVIDIA Jetson Xavier NX), the policy achieves an end-to-end inference latency below 1~ms.
Training follows a two-stage pipeline that initializes the policy through imitation learning and then fine-tunes it using reinforcement learning.
Across extensive evaluations, the resulting policy demonstrates effective racing performance.

The main contributions of this work are as follows:

\begin{enumerate}
\item We introduce End2Race, an end-to-end learning framework for autonomous racing that automatically generates overtaking scenarios at scale for policy training and evaluation.

\item We design a computationally efficient recurrent policy network with an inference latency below 1~ms, supporting real-time operation at high racing speeds.

\item Our learned policy generalizes to novel tracks and unseen opponent behaviors, showing high racing speeds, robust safety, and adaptive overtaking maneuvers.
\end{enumerate}

\begin{figure*}[!t]
    \centering
    \vspace{0.15cm}
    \includegraphics[width=\textwidth]{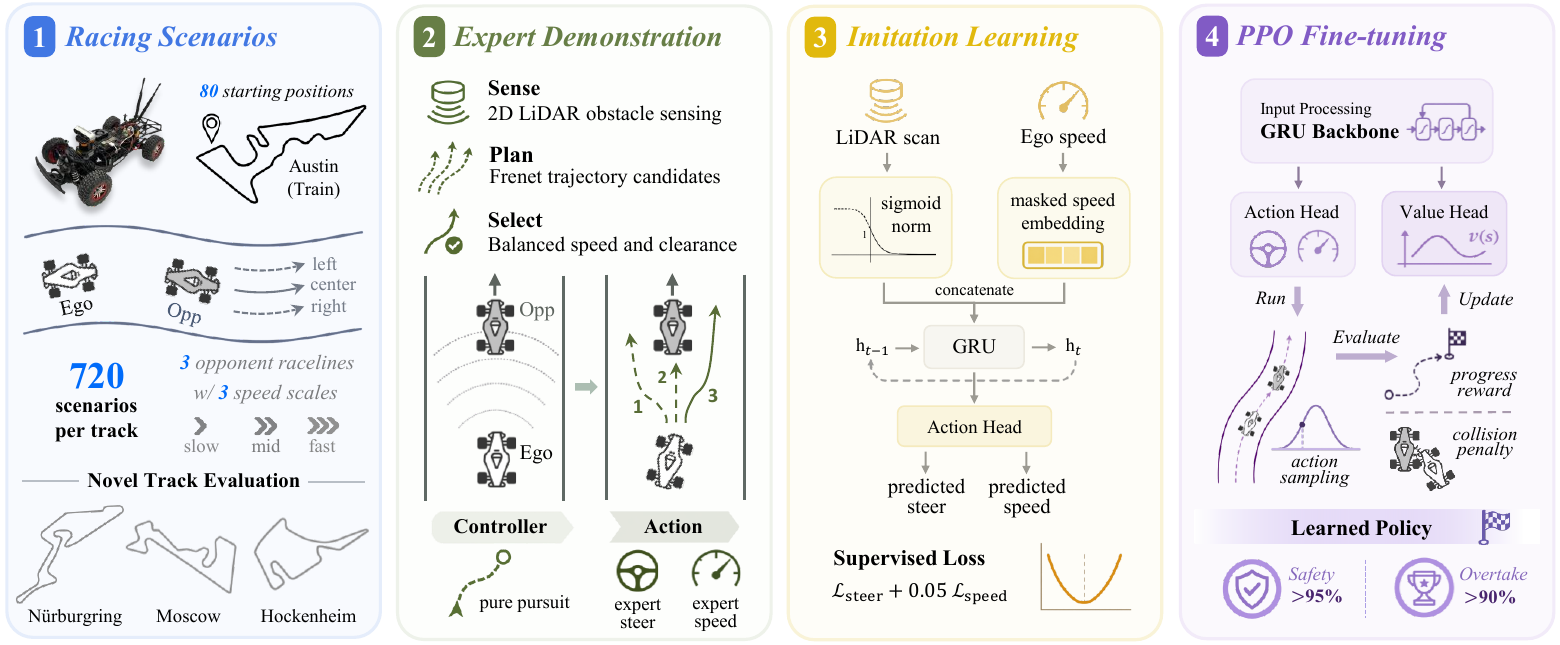}
    \caption{End2Race architecture.} 
    \label{fig:overview}
\end{figure*}

\section{Related Work}

\subsection{Imitation Learning}

Imitation learning (IL) trains a driving policy from expert demonstrations.
Sun et al.~\cite{10186780} generated training data using Pure Pursuit and evaluated Behavioral Cloning (BC)~\cite{bc}, DAgger~\cite{dagger}, HG-DAgger~\cite{hgdagger}, and EIL~\cite{eil}. They found that all policies completed the training track but failed to generalize to novel tracks.
Paluch et al.~\cite{paluch2024fpga} trained a feedforward policy to imitate nonlinear MPC along a raceline, achieving faster lap times than Pure Pursuit.
TinyLidarNet~\cite{zarrar2024tinylidarnet} trained a compact convolutional network to avoid static obstacles and evaluated it against dynamic opponents.
Zhang and Loidl~\cite{10275226} trained driving policies from human overtaking demonstrations, but relied on limited training and evaluation scenarios.

\subsection{Model-Free Reinforcement Learning}

Model-free RL optimizes control policies through environmental interaction to maximize cumulative reward.
Hamilton et al.~\cite{9763640} compared DDPG~\cite{ddpg}, SAC~\cite{sac}, and BC~\cite{bc}, finding that RL policies achieved faster lap times but incurred higher collision rates.
Steiner et al.~\cite{steiner2025overtaking} trained a TD3~\cite{td3} policy for head-to-head racing, demonstrating that opponent-inclusive training improves overtaking success. However, their deep feedforward network incurs substantial inference latency, which compromises real-time execution. Additionally, the policy is restricted to speeds below 2~m/s and fails above this limit.

\subsection{Model-Based Reinforcement Learning}

Model-based RL learns a predictive model of the environment and uses it to optimize control policies through imagined rollouts.
For instance, Dreamer~\cite{hafner2019dream} learns a recurrent state-space model and trains actor-critic networks entirely within a latent space.
Brunnbauer et al.~\cite{9811650} applied Dreamer to autonomous racing, reporting better sample efficiency and lap completion than model-free methods.

\subsection{Hybrid Learning Methods}

Hybrid methods integrate learning-based modules into classical planning or control pipelines.
For example, Dwivedi et al.~\cite{10003698} combined high-level path guidance from a global planner with a low-level learned tracking controller.
To guide overtaking decisions, Bhargav et al.~\cite{bhargav2021offline} modeled overtaking success probabilities across track segments to dynamically select lateral planning modes.
Similarly, Trumpp et al.~\cite{10801657} used an artificial potential field planner for baseline obstacle avoidance and trained a residual RL policy to provide corrective control for overtaking.

\section{Methodology}

We develop End2Race using the F1Tenth simulator, which provides high-fidelity vehicle dynamics and accurate 2D LiDAR scans.
The simulated sensor produces 360-degree scans of 1440 range beams with a 10~m maximum range at 40~Hz.
Control commands follow the platform's physical limits, with steering angles bounded within $\pm 0.42\,\text{rad}$ and speed up to $17.9\,\text{m/s}$.
Four tracks are selected: Austin, Hockenheim, MoscowRaceway, and Nuerburgring, each reflecting a real-world track layout.
Among these, Austin is used for training, while the rest are held out for evaluation.
For each track, an automated workflow generates diverse overtaking scenarios that pair an ego vehicle with an opponent.
Using demonstrations from a map-based expert, we first train a GRU-based network through BC, and then fine-tune the resulting policy with Proximal Policy Optimization (PPO)~\cite{ppo}.
The following sections describe each stage in detail (Fig.~\ref{fig:overview}).

\subsection{Overtaking Scenario Setup}

Each overtaking scenario pairs an ego vehicle with an opponent initially positioned ahead of it and has a duration of 8 seconds. Vehicle placement is expressed in Frenet coordinates, in which the longitudinal direction measures progress along the track and the lateral direction denotes position across the track width. We specify each scenario using four parameters, \(\left(s_0,\,\Delta s_0,\,r_{\mathrm{opp}},\,\alpha_{\mathrm{opp}}\right)\).

\subsubsection{Ego Starting Position (\(s_0\))}
For each track, we define 80 ego starting positions equally spaced along the track. On the 420~m Austin track, for example, adjacent starting positions are separated by approximately 5.2~m.

\subsubsection{Longitudinal Gap (\(\Delta s_0\))}
The parameter \(\Delta s_0\) specifies how far the opponent is positioned ahead of the ego vehicle at the start of a scenario. A small gap provides insufficient maneuvering space, while a large gap requires more time to catch the opponent, leaving less time for overtaking. We therefore set \(\Delta s_0=3\)~m to balance these considerations.

\subsubsection{Opponent Raceline (\(r_{\mathrm{opp}}\))}
Beyond longitudinal placement, the lateral positions of the ego vehicle and opponent are critical to the overtaking outcome. We represent this configuration using the opponent raceline \(r_{\mathrm{opp}}\), which assigns the opponent to one of three reference paths. Using the raceline generation method of Heilmeier et al.~\cite{Heilmeier02102020}, we define left, center, and right paths positioned at \(35\%\), \(50\%\), and \(65\%\) across the track width from the left boundary. On the Austin track, this yields approximately \(0.25\)~m of lateral separation between adjacent racelines. An opponent on the center raceline restricts clearance on both sides, whereas one on a side raceline leaves a wider passing corridor.

\subsubsection{Opponent Speed Scale (\(\alpha_{\mathrm{opp}}\))}
Each reference raceline provides an optimized speed profile based on track curvature and vehicle dynamics. Here, we set a maximum speed of 7.5~m/s, as empirical testing showed that higher speeds compromise the tracking stability of the rule-based controller. If the opponent follows the raceline speed profile exactly, an ego vehicle operating under the same profile cannot catch up. To create overtaking opportunities, we multiply the opponent's reference speed by a factor \(\alpha_{\mathrm{opp}}\in\{0.4,0.6,0.8\}\). Values below 0.4 make the opponent unrealistically slow, while those above 0.8 often prevent the ego vehicle from catching up in time.

Combining 80 ego starting positions, a fixed longitudinal gap, three opponent racelines, and three opponent speed scales yields 720 scenarios per track. During each 8-second scenario, the opponent tracks its assigned reference raceline and speed profile using a Pure Pursuit controller. It does not engage in adversarial behaviors such as active blocking or erratic speed variations~\cite{zheng2025bridginggapdiscreteagent}. This setup aligns with the current state of the field, where overtaking and collision avoidance remain open challenges before defensive tactics can be explored.

\subsection{Expert Demonstration}

We use a map-based lattice planner to generate overtaking demonstrations for the ego vehicle. Here, the ego vehicle uses the center raceline as its global reference path and operates under the full speed profile. The lattice planner, adapted from the Frenet optimal trajectory framework~\cite{sakai2018pythonrobotics}, generates local candidate trajectories across a range of lateral offsets and terminal speeds relative to this reference. This allows the ego vehicle to temporarily deviate from the centerline to avoid obstacles and execute overtaking maneuvers.

At each timestep, the lattice planner generates a set of candidate trajectories around the reference path, discards those that violate vehicle dynamic constraints, and selects the trajectory \(\tau\) that minimizes the cost function \(\mathcal{J}(\tau)\):
\begin{equation}
    \mathcal{J}(\tau)
    =
    \frac{1}{N}\sum_{i=1}^{N}
    \left(1-\frac{v_i}{v_{\max}}\right)
    +
    \lambda_c
    \max_{i}
    \left[
        \max\left(0,1-\frac{d_i}{d_c}\right)
    \right]^2
\end{equation}
where \(N=6\) is the number of future trajectory points sampled at 0.1~s intervals; \(v_i\) is the projected vehicle speed at point \(i\); \(v_{\max}=7.5\)~m/s is the speed limit; \(d_i\) is the distance from point \(i\) to the nearest obstacle detected by the LiDAR scan; and \(d_c=0.5\)~m is the clearance threshold. The first term encourages the vehicle to travel at higher speeds up to the speed limit, while the second term penalizes proximity to obstacles. They are balanced by the collision weight \(\lambda_c=5\), determined empirically. The selected local trajectory is then tracked using Pure Pursuit.

\begin{figure}[!t]
    \centering
    \vspace{0.15cm}
    \setlength{\fboxsep}{0pt}
    \setlength{\fboxrule}{0.2pt}
    \begin{tabular}{@{}c@{\hspace{0.8mm}}c@{\hspace{0.8mm}}c@{}}
        \fcolorbox{black!10}{white}{\includegraphics[width=0.315\linewidth]{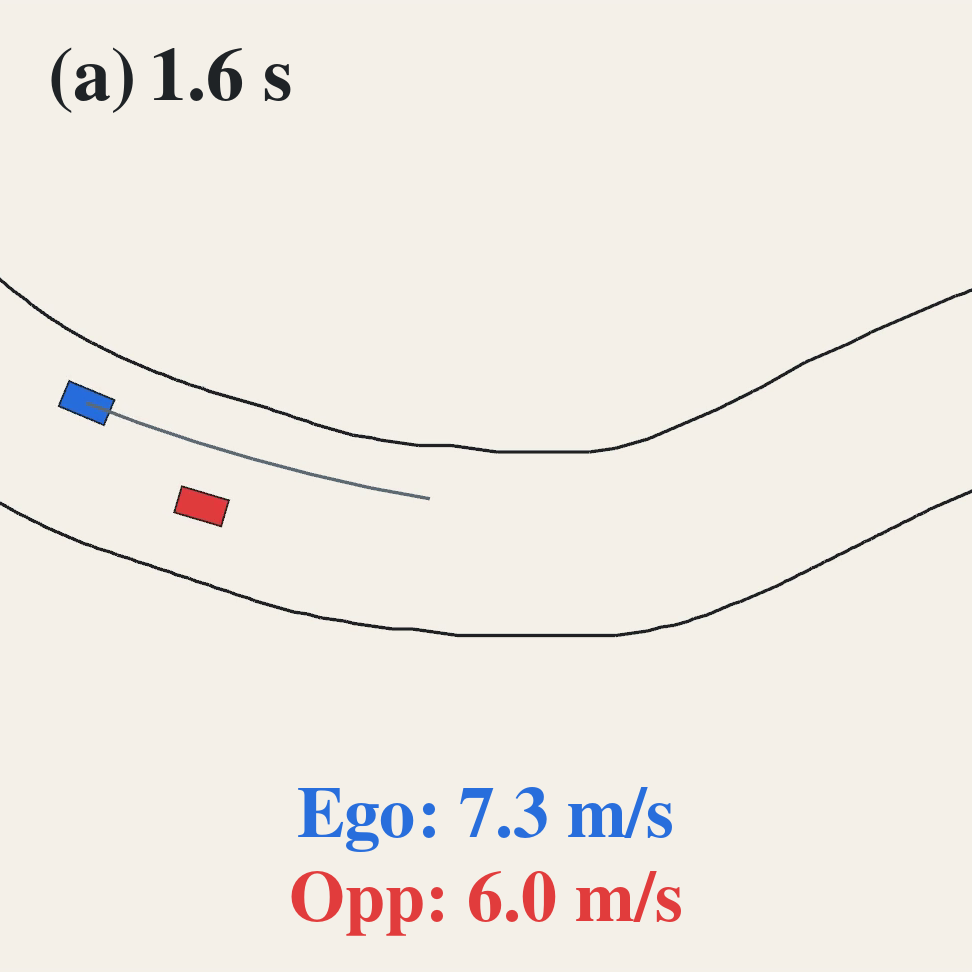}} &
        \fcolorbox{black!10}{white}{\includegraphics[width=0.315\linewidth]{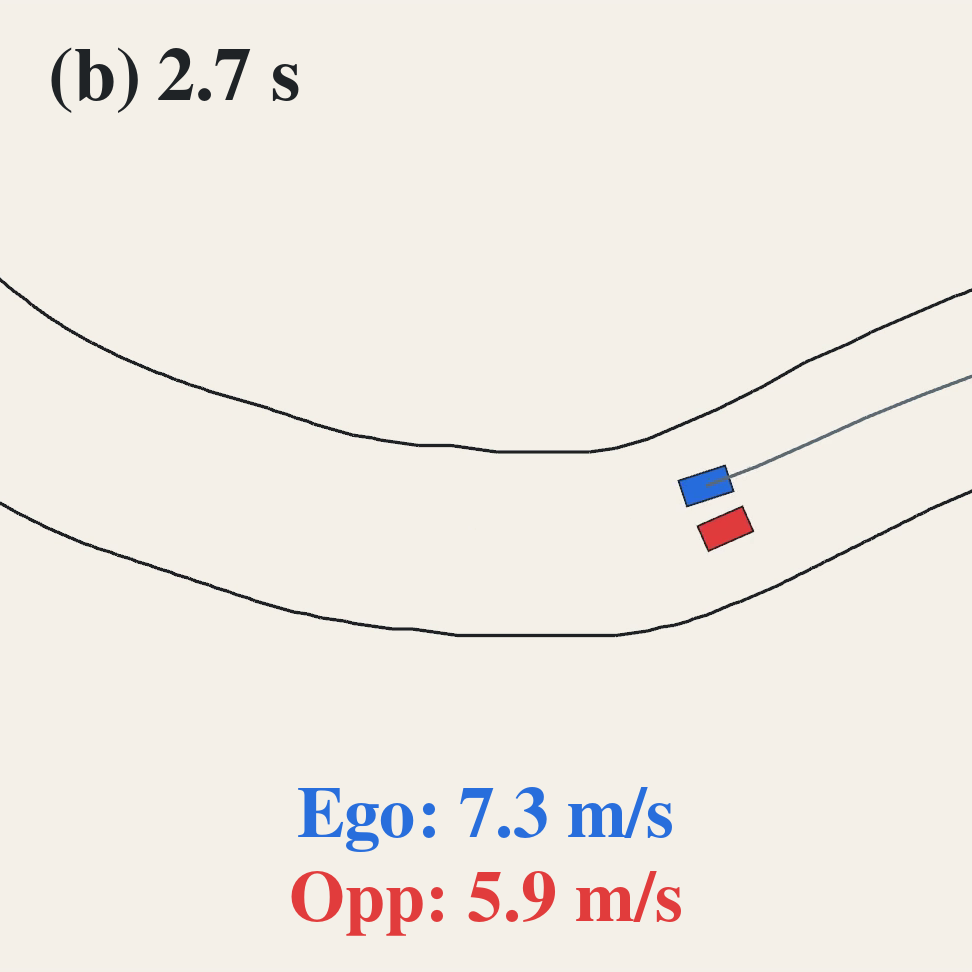}} &
        \fcolorbox{black!10}{white}{\includegraphics[width=0.315\linewidth]{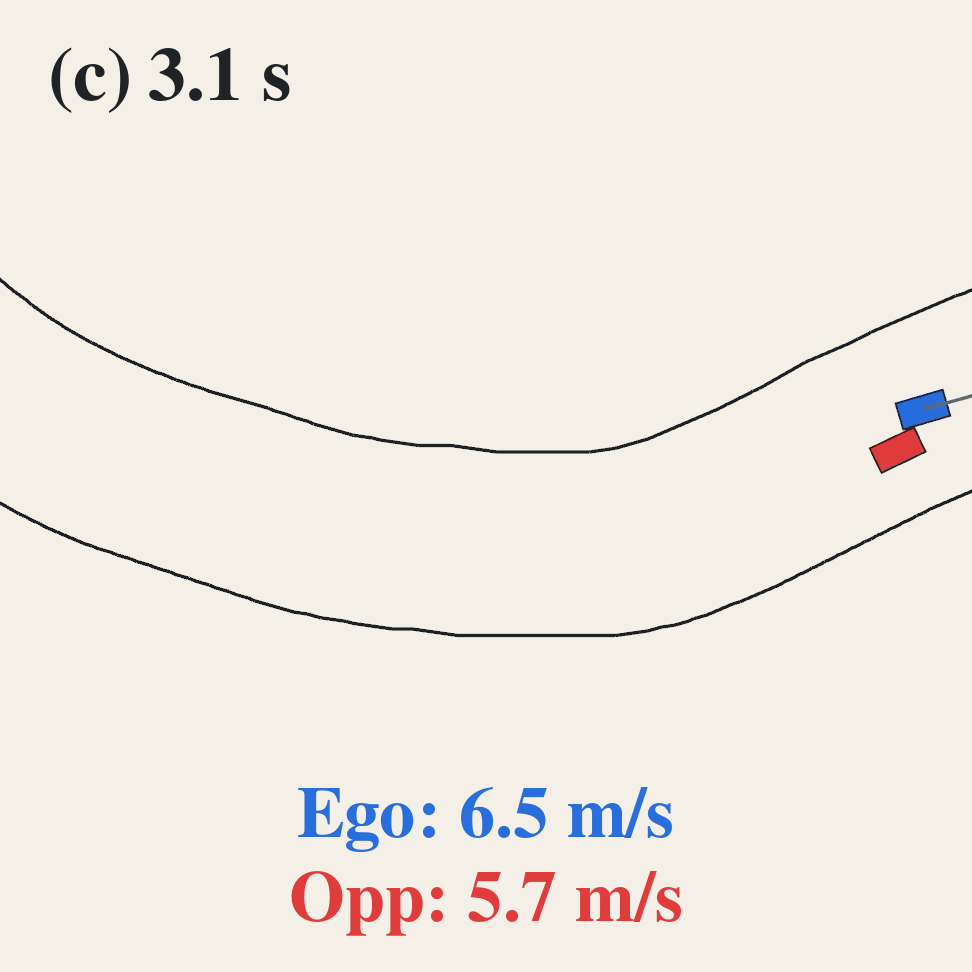}}
    \end{tabular}
    \caption{Expert-planner failure during overtaking, shown in temporal order from (a) to (c). Blue denotes the ego vehicle and its planned trajectory, and red denotes the opponent. In this scenario, the opponent follows the right raceline with a speed scale of 0.8.}
    \label{fig:expert-failure}
\end{figure}

\subsection{Training Dataset}
\label{sec:training-dataset}

We collect 720 eight-second expert demonstrations on the Austin track. Training data is recorded at 40~Hz to match the onboard LiDAR frequency, yielding 320 observation-action pairs per scenario. Of these, 620 result in successful overtakes, 23 in car-following, and 77 in collisions. An overtake is counted when the ego vehicle advances at least one vehicle length ($0.58\,\text{m}$) ahead of the opponent, while collisions include contact with either the opponent or the track boundary. The expert collisions stem primarily from the lattice planner's forward-only obstacle avoidance, which does not account for a moving opponent alongside the ego vehicle. Consequently, the ego vehicle may return to the center before it has fully cleared the opponent, causing a sideswipe collision. This risk increases as the speed difference between the two vehicles decreases, with most failures occurring at an opponent speed scale of 0.8. Figure~\ref{fig:expert-failure} illustrates one such failure, where the ego vehicle prematurely returns to the centerline and collides with the opponent.

Such failures can be mitigated with explicit opponent detection, trajectory prediction, and interaction modeling, as shown in Predictive Spliner~\cite{baumann2025predictive}. However, their implementation requires following the opponent for an entire lap before attempting a pass, which does not reflect how vehicles race in practice. We instead exclude the collision demonstrations from BC training and rely on downstream PPO for policy refinement (Sec.~\ref{sec:policy-training}).

\subsection{Model Architecture}

We use a GRU-based network to capture temporal dependencies and map sequential observations to vehicle control commands.
At each timestep, the observation consists of a 2D LiDAR scan and the current ego speed.
To reduce computation, we uniformly downsample the raw 1440 beams to 180.
The control output specifies the desired speed and steering angle.
The following introduces key design details, including LiDAR normalization, ego-speed projection, and the network design.

\subsubsection{LiDAR Normalization}

Bufacchi and Iannetti show that human interaction with surrounding space is proximity-dependent, with behavioral responses strengthening progressively as object distance decreases~\cite{bufacchi2018action}.
The same principle can be applied to autonomous racing, where the influence of surrounding obstacles on ego motion similarly depends on their distance.
To emulate this aspect of human spatial cognition, we transform each of the 180 LiDAR beams using a sigmoid function that maps the raw scan value, \(x \in [0, 10]\), to a normalized value, \(\sigma(x) \in [0,1]\):
\begin{equation}
    \sigma(x) = \left(-\frac{1}{1 + e^{-k x}} + 1\right) \cdot 2
\end{equation}
Here, the parameter $k$ determines the decay rate across the sensing range. We set $k=0.3$ to prioritize immediate obstacles within $0\text{--}5$~m, while maintaining reduced sensitivity for distant geometry.

\subsubsection{Ego-Speed Projection}

To maintain dynamic consistency, we incorporate the current ego speed as a policy input, projecting the scalar speed \(v_t\) through a linear layer into a 30-dimensional vector \(\psi(v_t)\).
To prevent shortcut learning~\cite{shortcut}, in which the model simply copies the current speed into its prediction, we randomly mask the projected speed with a learnable embedding $\mathbf{e}_{\mathrm{mask}}$ as follows:
\begin{equation}
\widetilde{\psi}(v_t) =
\begin{cases}
    \psi(v_t), & \text{with probability } 1-p, \\
    \mathbf{e}_{\mathrm{mask}}, & \text{with probability } p,
\end{cases}
\end{equation}
Here, we set \(p=0.2\), encouraging the model to determine desired speed from LiDAR observations when the current speed is masked and to incorporate it when available.

\subsubsection{Network Design}

The 180-dimensional normalized LiDAR scan and the 30-dimensional speed projection are concatenated to form a 210-dimensional input vector.
This input is passed to a single-layer GRU network with a 420-dimensional hidden state.
The hidden state is initialized to zero at the start of each episode and updated at each timestep.
It is then passed to a two-layer MLP with ReLU activation to produce the desired speed and steering angle.
In total, the model has 850,556 trainable parameters.

\begin{figure}[!t]
    \centering
    \vspace{0.15cm}
    \begin{minipage}[t]{0.49\columnwidth}
        \centering
        \includegraphics[width=\linewidth]{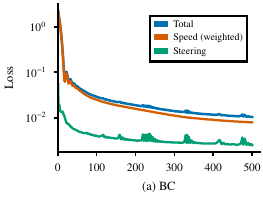}
    \end{minipage}%
    \hfill
    \begin{minipage}[t]{0.49\columnwidth}
        \centering
        \includegraphics[width=\linewidth]{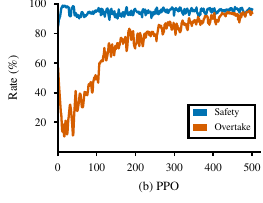}
    \end{minipage}
    \vspace{-1.5mm}
    \caption{Training curves for the two-stage policy optimization, with each stage trained for 500 epochs. (a) BC loss, with the speed loss weighted by 0.05. (b) PPO safety and overtake rates on the 720 Austin scenarios.}
    \label{fig:training-curves}
\end{figure}

\begin{table*}[!t]
\vspace{0.15cm}
\caption{Single-Vehicle Timed Trials Performance}
\centering
\small
\setlength{\tabcolsep}{1.8pt}
\begin{tabularx}{\textwidth}{>{\raggedright\arraybackslash}p{2.8cm}|YY|YY|YY|YY}
\hline
\multirow{2}{*}{\textbf{Method}} &
\multicolumn{2}{c|}{\textbf{Austin (Train)}} &
\multicolumn{2}{c|}{\textbf{Hockenheim}} &
\multicolumn{2}{c|}{\textbf{MoscowRaceway}} &
\multicolumn{2}{c}{\textbf{Nuerburgring}} \\[-0.65ex]
& Time (s) & Speed (m/s) & Time (s) & Speed (m/s) & Time (s) & Speed (m/s) & Time (s) & Speed (m/s) \\
\hline
Dreamer~\cite{9811650} & 140.2 & 3.0 & 118.2 & 3.0 & 112.8 & 2.8 & 141.0 & 3.1 \\
TinyLidarNet~\cite{zarrar2024tinylidarnet} & 79.8 & 5.2 & 65.8 & 5.5 & 59.8 & 5.3 & 78.2 & 5.7 \\
Zhang--Loidl~\cite{10275226} & -- & -- & -- & -- & -- & -- & 66.4 & 6.6 \\
Steiner~\cite{steiner2025overtaking} & 207.2 & 2.0 & 177.0 & 2.0 & 157.0 & 2.0 & 218.4 & 2.0 \\
Pred. Spliner~\cite{baumann2025predictive} & 61.7 & 6.7 & 49.9 & 7.1 & 48.8 & 6.5 & 60.7 & 7.3 \\
End2Race (Expert) & 63.2 & 6.7 & 53.2 & 6.8 & 50.6 & 6.5 & 64.7 & 6.9 \\
End2Race (BC) & 62.7 & 6.7 & 52.5 & 6.9 & 47.7 & 6.8 & 63.7 & 7.0 \\
\textbf{End2Race (PPO)} & \textbf{48.7} & \textbf{8.7} & \textbf{41.1} & \textbf{8.8} & \textbf{38.5} & \textbf{8.5} & \textbf{50.2} & \textbf{8.9} \\
\hline
\end{tabularx}
\label{tab:single_vehicle}
\par\vspace{2.0mm}
\caption{Head-to-Head Racing Performance}
\centering
\small
\setlength{\tabcolsep}{1.8pt}
\begin{tabularx}{\textwidth}{>{\raggedright\arraybackslash}p{2.8cm}|YY|YY|YY|YY}
\hline
\multirow{2}{*}{\textbf{Method}} &
\multicolumn{2}{c|}{\textbf{Austin (Train)}} &
\multicolumn{2}{c|}{\textbf{Hockenheim}} &
\multicolumn{2}{c|}{\textbf{MoscowRaceway}} &
\multicolumn{2}{c}{\textbf{Nuerburgring}} \\[-0.65ex]
& Safety (\%) & Overtake (\%) & Safety (\%) & Overtake (\%) & Safety (\%) & Overtake (\%) & Safety (\%) & Overtake (\%) \\
\hline
TinyLidarNet~\cite{zarrar2024tinylidarnet} & 80.3 & 28.9 & 81.3 & 39.3 & 85.8 & 46.4 & 78.6 & 37.5 \\
End2Race (Expert) & 89.3 & 86.1 & 91.8 & 90.6 & 87.2 & 83.5 & 92.8 & 92.4 \\
End2Race (BC) & 82.8 & 62.8 & 78.6 & 66.1 & 76.7 & 69.0 & 81.5 & 74.6 \\
\textbf{End2Race (PPO)} & \textbf{98.1} & \textbf{96.1} & \textbf{97.5} & \textbf{97.1} & \textbf{97.1} & \textbf{96.8} & \textbf{97.9} & \textbf{97.5} \\
\hline
\end{tabularx}
\label{tab:head_to_head}
\end{table*}

\subsection{Policy Optimization}
\label{sec:policy-training}

Training consists of two stages: we first use BC to learn a policy from expert demonstrations, and then fine-tune it with PPO via direct environment interaction.

\subsubsection{Behavioral Cloning}
\label{sec:behavioral-cloning}

Each 8-second scenario is processed as a whole, generating a control prediction at every timestep and aggregating the loss across the full sequence. The loss function combines the mean squared error (MSE) of speed (m/s) and steering angle (rad), weighting the speed loss by 0.05 to balance their scales:
\begin{equation}
\mathcal{L} = 0.05\, \mathcal{L}_{\text{speed}} + \mathcal{L}_{\text{steer}}.
\end{equation}

\subsubsection{Proximal Policy Optimization}
\label{sec:rl-finetuning}

The policy is fine-tuned across the 720 scenarios on the Austin track. In each training iteration, we collect one trajectory per scenario, combining them into a single batch to perform a policy update. During rollout, actions are sampled from a Gaussian distribution centered at the network outputs. At each timestep $t$, the reward function $r_t$ combines a raceline-progress term with a collision penalty:
\begin{equation}
r_t = 0.01d_t - 3c_t,
\end{equation}
where \(d_t\) denotes the distance traveled along the raceline, \(c_t\) is a binary collision indicator, and the coefficients are chosen empirically. Together, these terms encourage the ego vehicle to pursue high racing speeds while prioritizing safety during overtaking maneuvers.

\section{Experiments}
\label{sec:experiments}

\begin{figure*}[!t]
    \centering
    \vspace{0.15cm}
    \setlength{\fboxsep}{0pt}
    \setlength{\fboxrule}{0.2pt}
    \begin{minipage}[t]{0.485\textwidth}
        \centering
        \fcolorbox{black!10}{white}{\includegraphics[width=0.322\linewidth]{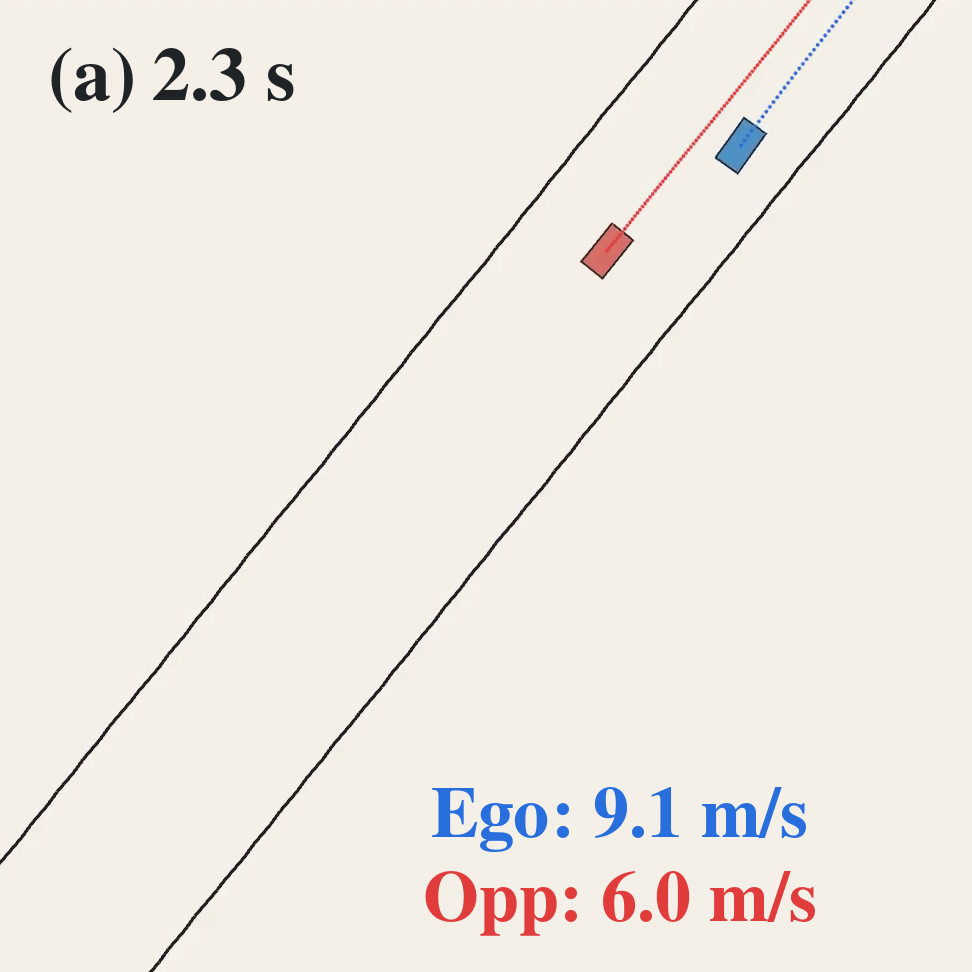}}%
        \hfill
        \fcolorbox{black!10}{white}{\includegraphics[width=0.322\linewidth]{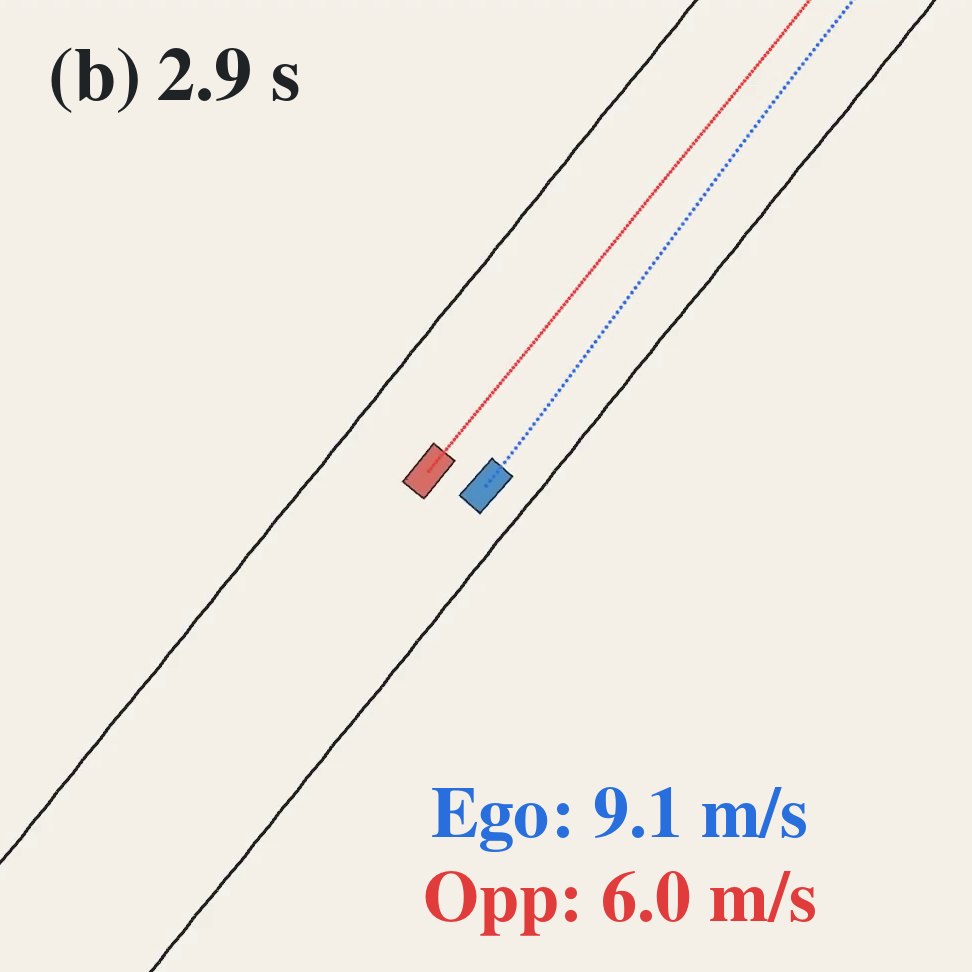}}%
        \hfill
        \fcolorbox{black!10}{white}{\includegraphics[width=0.322\linewidth]{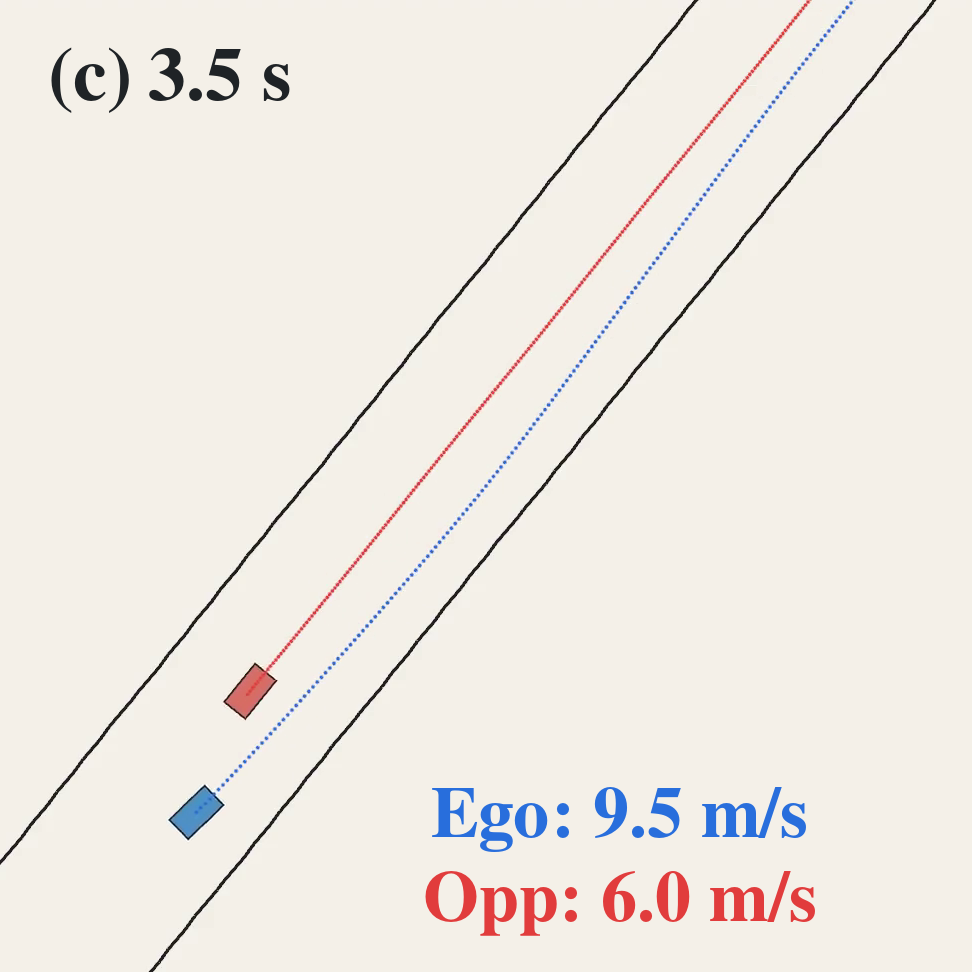}}
        \par\vspace{0.4mm}{\footnotesize (a) Straight overtake ($r_{\mathrm{opp}}=\text{center},\, \alpha_{\mathrm{opp}}=0.8$)}
    \end{minipage}%
    \hfill
    \begin{minipage}[t]{0.485\textwidth}
        \centering
        \fcolorbox{black!10}{white}{\includegraphics[width=0.322\linewidth]{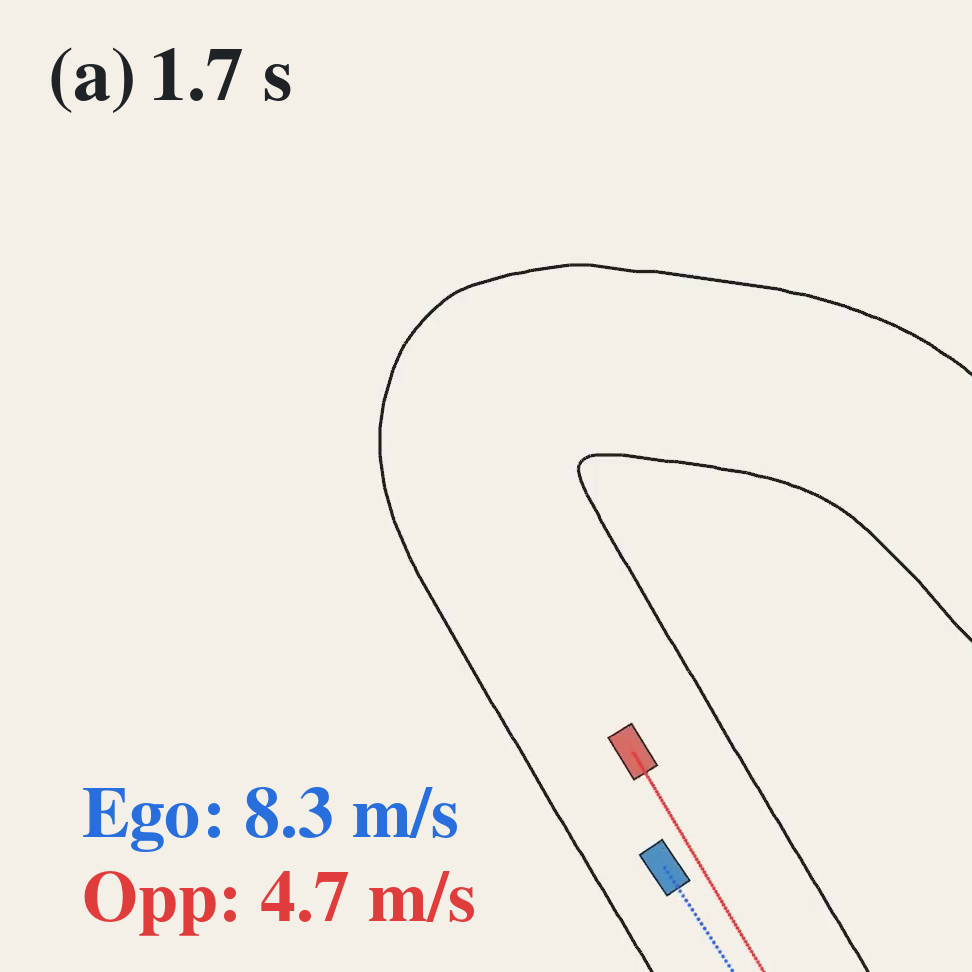}}%
        \hfill
        \fcolorbox{black!10}{white}{\includegraphics[width=0.322\linewidth]{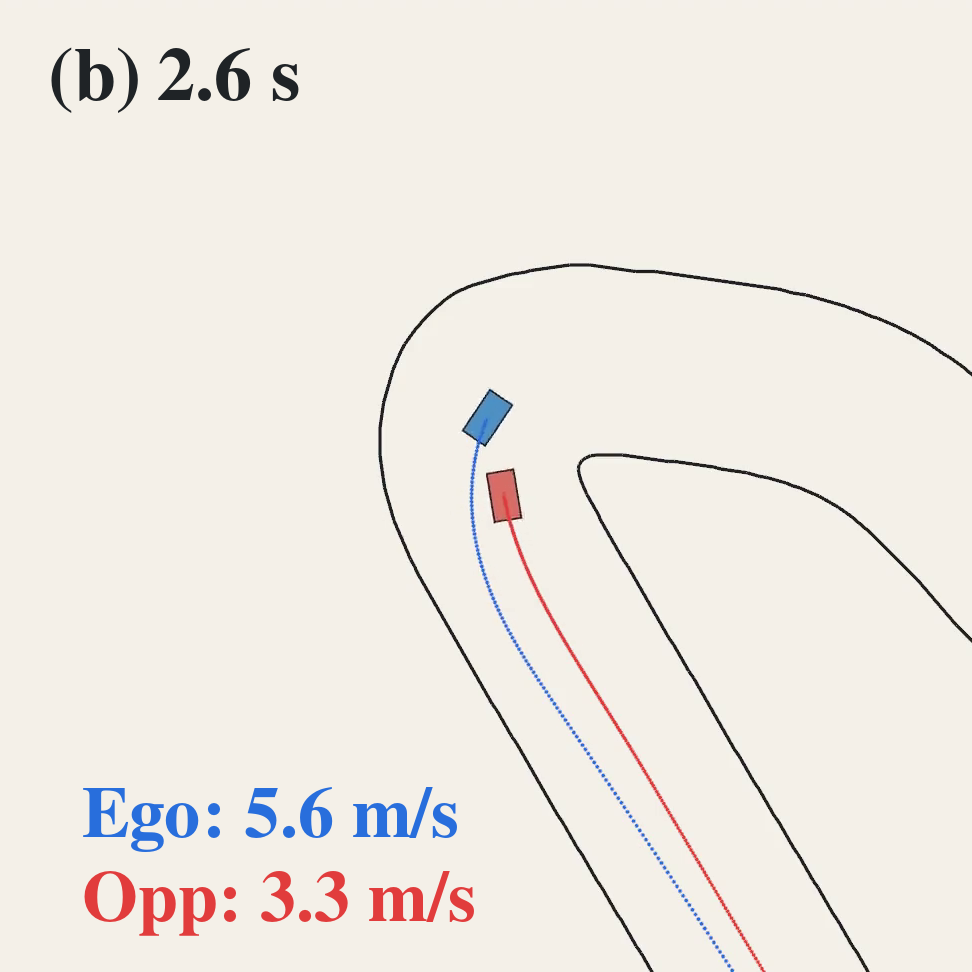}}%
        \hfill
        \fcolorbox{black!10}{white}{\includegraphics[width=0.322\linewidth]{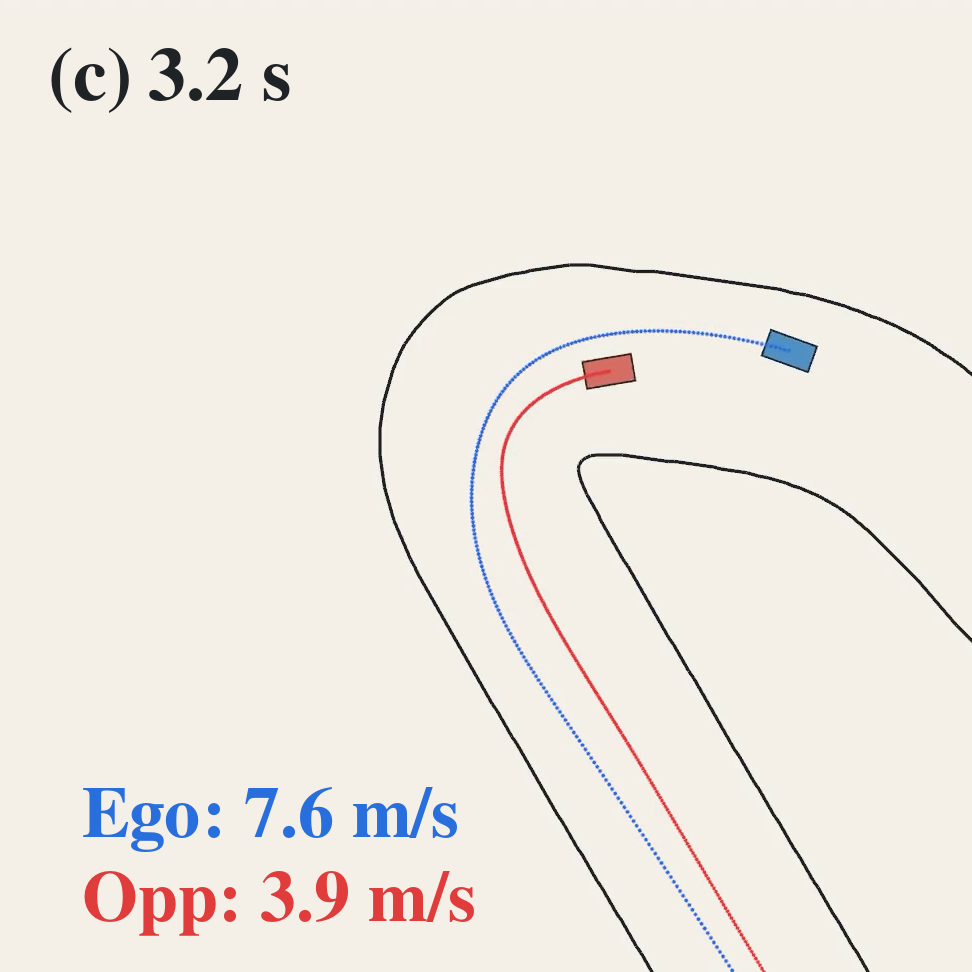}}
        \par\vspace{0.4mm}{\footnotesize (b) Cornering overtake ($r_{\mathrm{opp}}=\text{center},\, \alpha_{\mathrm{opp}}=0.8$)}
    \end{minipage}

    \vspace{1.5mm}

    \begin{minipage}[t]{0.485\textwidth}
        \centering
        \fcolorbox{black!10}{white}{\includegraphics[width=0.322\linewidth]{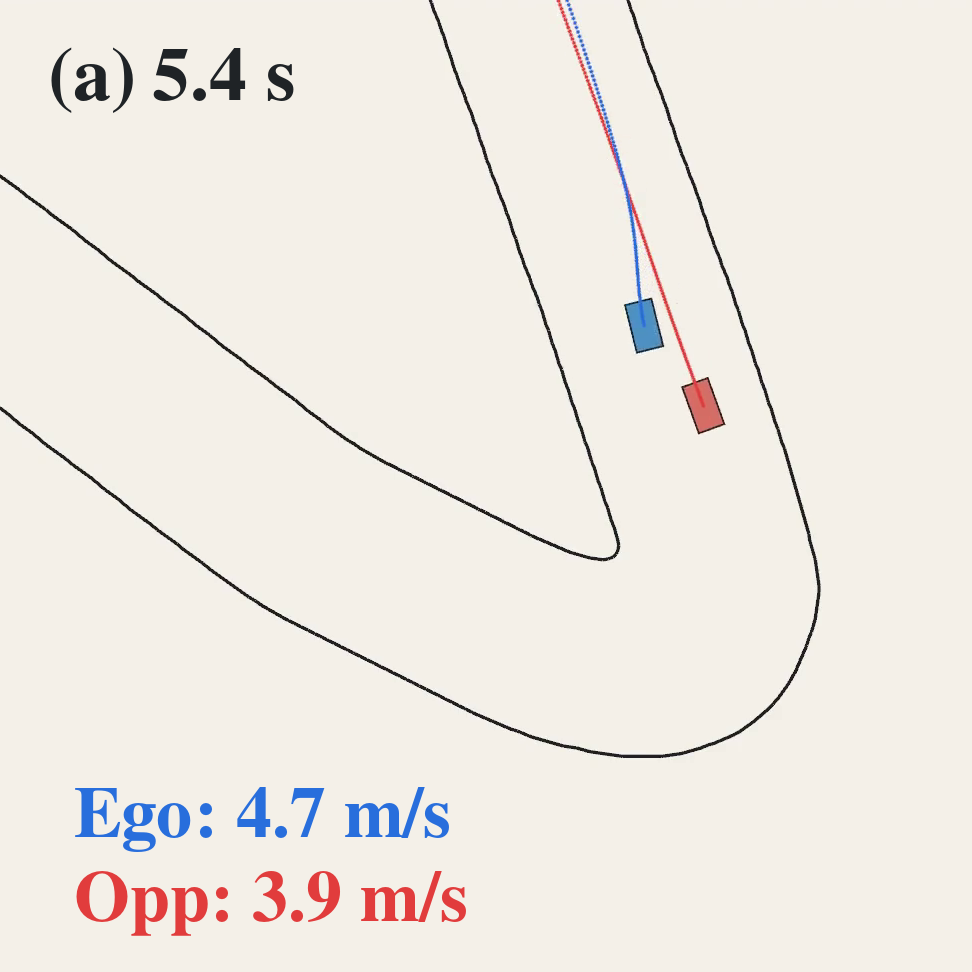}}%
        \hfill
        \fcolorbox{black!10}{white}{\includegraphics[width=0.322\linewidth]{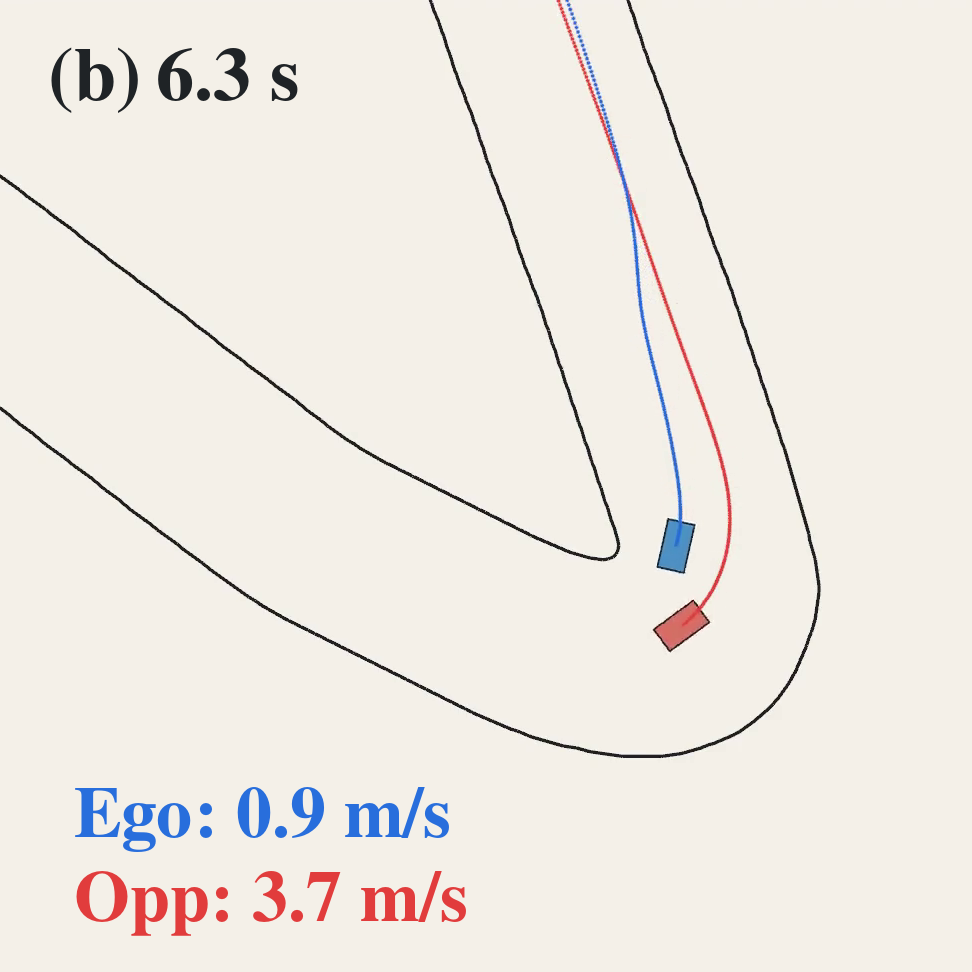}}%
        \hfill
        \fcolorbox{black!10}{white}{\includegraphics[width=0.322\linewidth]{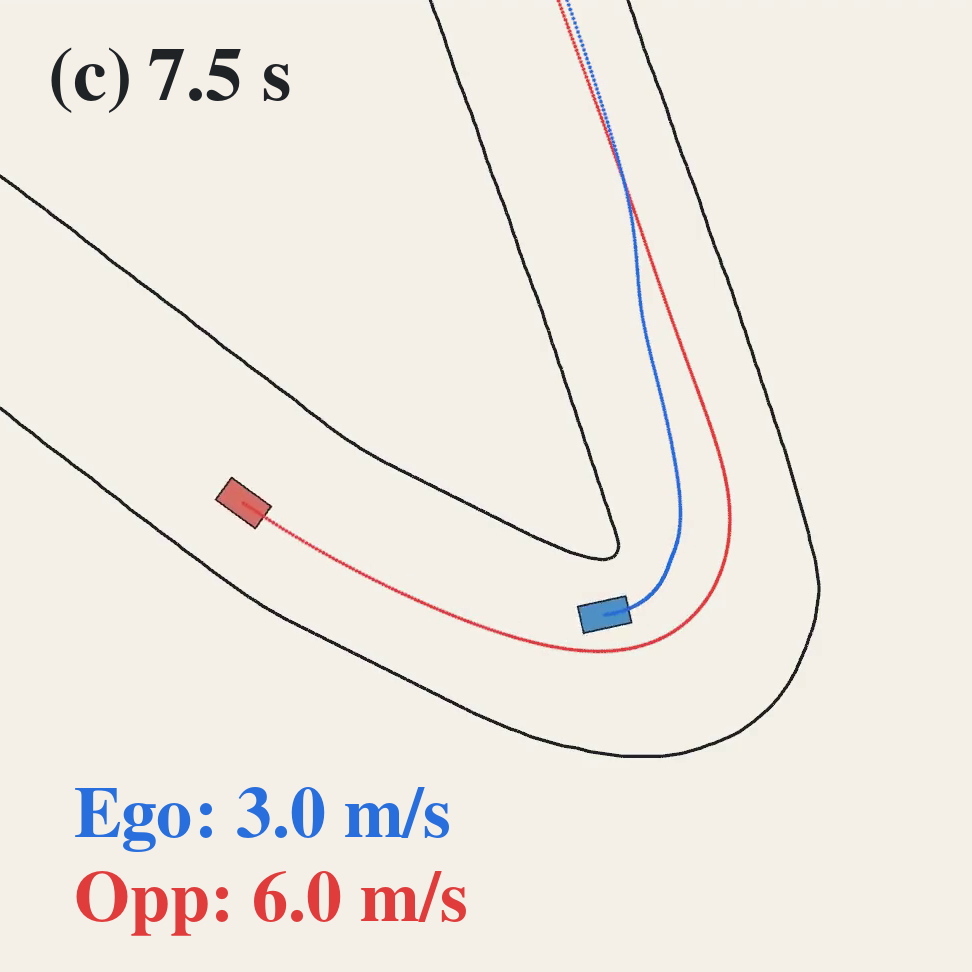}}
        \par\vspace{0.4mm}{\footnotesize (c) Car-following ($r_{\mathrm{opp}}=\text{right},\, \alpha_{\mathrm{opp}}=0.8$)}
    \end{minipage}%
    \hfill
    \begin{minipage}[t]{0.485\textwidth}
        \centering
        \fcolorbox{black!10}{white}{\includegraphics[width=0.322\linewidth]{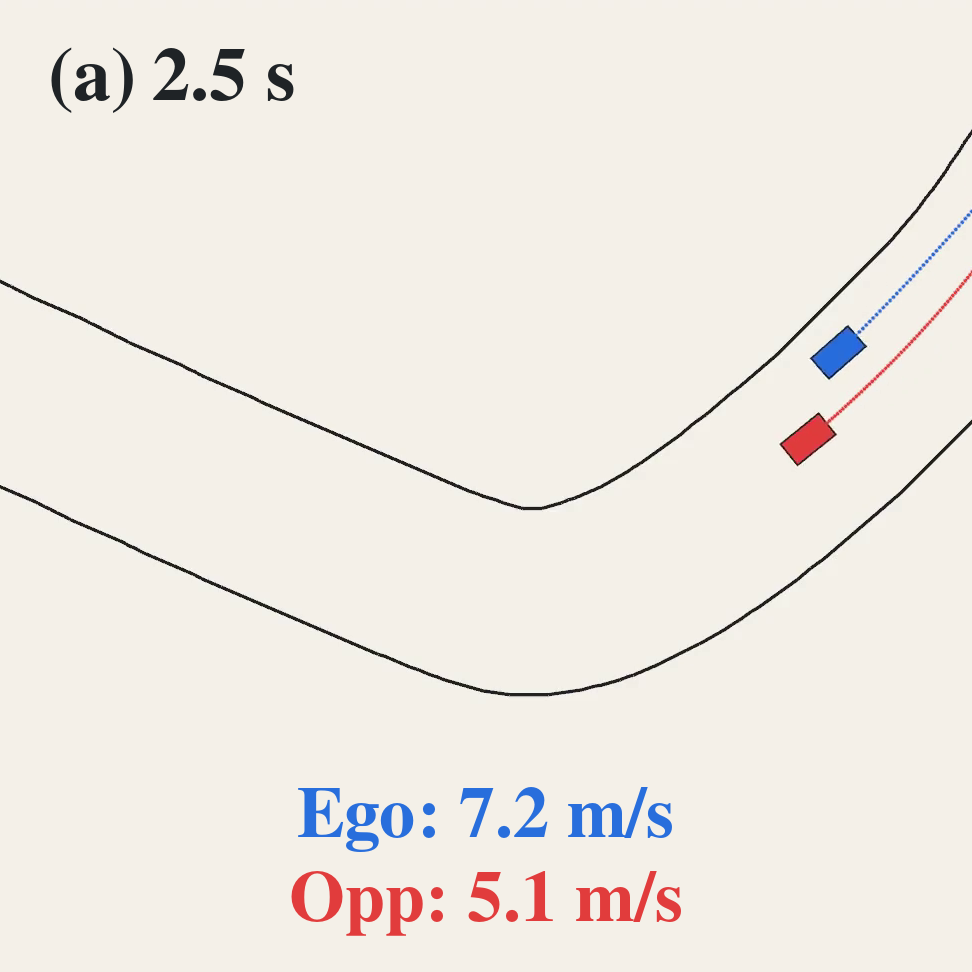}}%
        \hfill
        \fcolorbox{black!10}{white}{\includegraphics[width=0.322\linewidth]{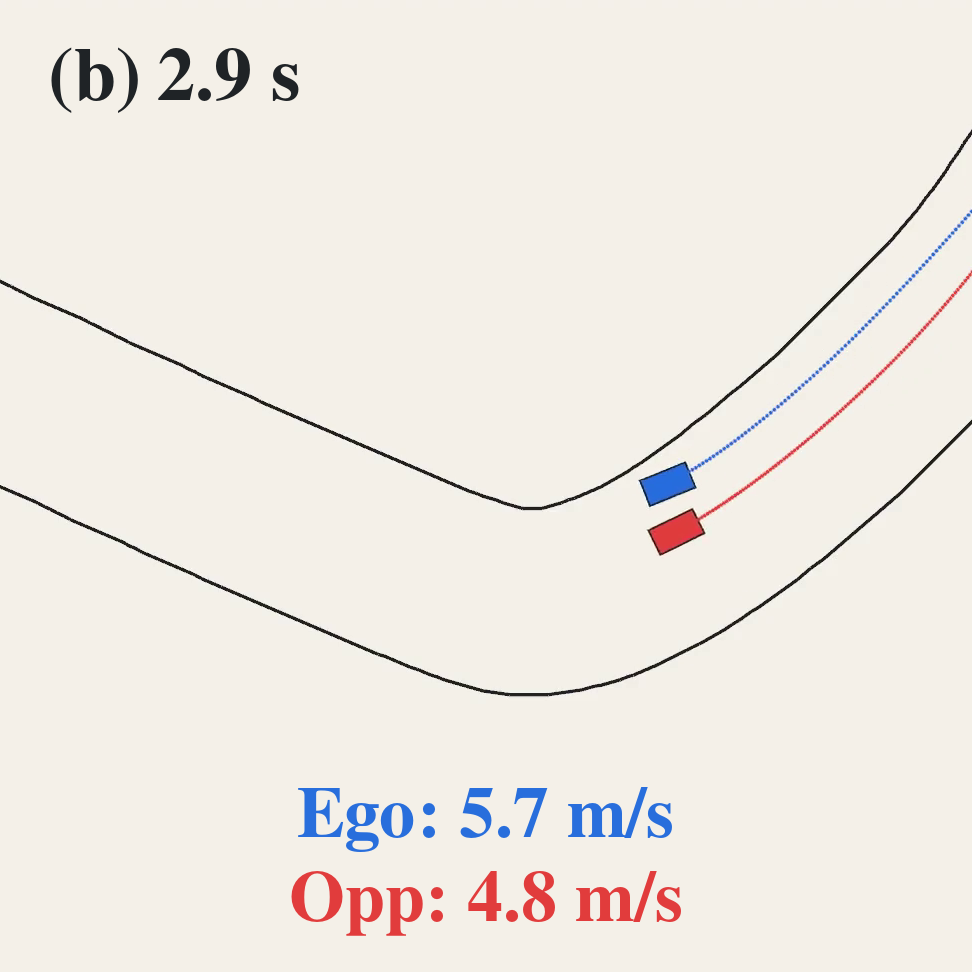}}%
        \hfill
        \fcolorbox{black!10}{white}{\includegraphics[width=0.322\linewidth]{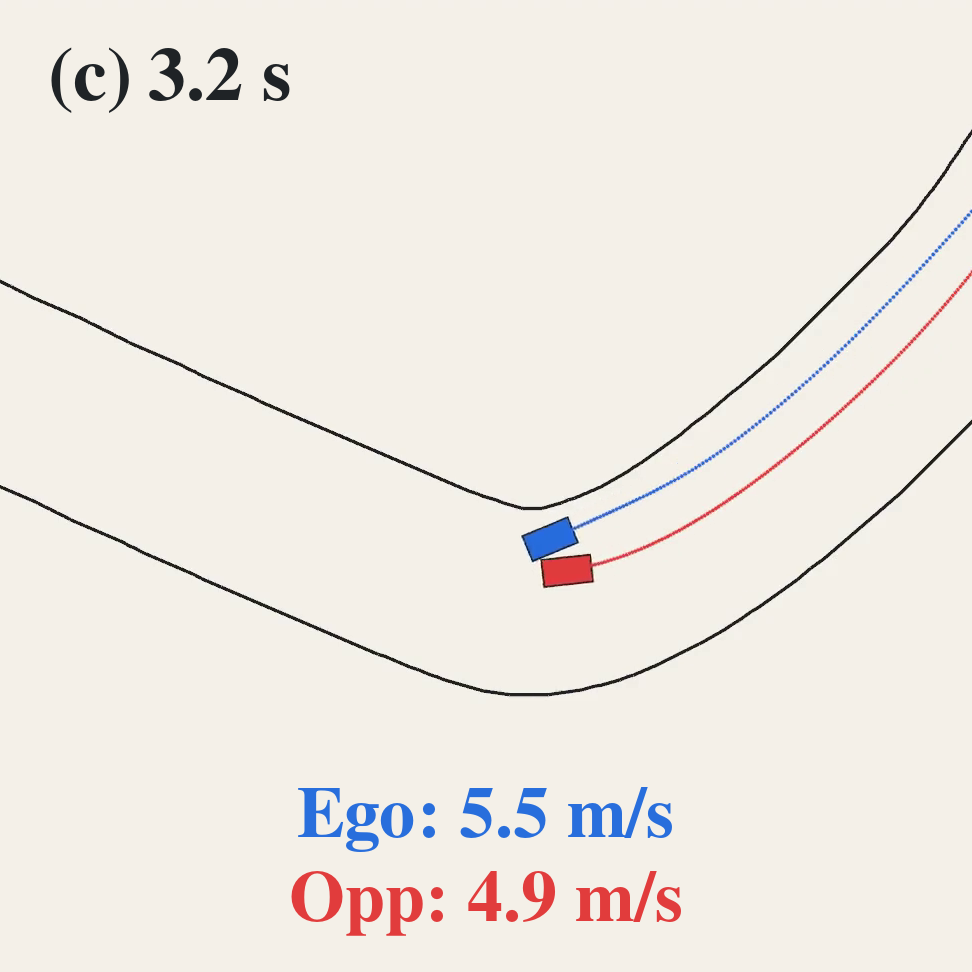}}
        \par\vspace{0.4mm}{\footnotesize (d) Collision ($r_{\mathrm{opp}}=\text{center},\, \alpha_{\mathrm{opp}}=0.8$)}
    \end{minipage}
    \caption{Head-to-head scenarios on the Austin track. Blue and red denote the ego vehicle and opponent, respectively, with their trajectory history rendered in matching colors.}
    \label{fig:scenario-examples}
\end{figure*}

We evaluate End2Race on the training track and three novel test tracks, comparing it against open-source learning-based methods from the literature. Each method is evaluated using its best released checkpoint; those lacking public code or checkpoints (e.g.,~\cite{10003698,bhargav2021offline,10801657}) are omitted. Experiments cover two setups: single-vehicle timed trials and head-to-head racing. In addition to performance metrics, trajectory analysis examines representative policy behaviors.

\subsection{Training Details}

The BC policy is trained for 500 epochs using the Adam optimizer with a learning rate of \(10^{-4}\). Training completes in five minutes on a single L40S GPU. The loss decreases rapidly during early epochs and then gradually converges. Additional training would further reduce the loss, but risks overfitting without yielding performance gains (Fig.~\ref{fig:training-curves}(a)).

PPO starts from the BC checkpoint with a newly initialized value head. Training runs for 500 epochs with a learning rate of \(10^{-5}\) and a clipping range of \([0.8, 1.2]\). Policy updates use GAE with discount factor $\gamma = 0.99$ and $\lambda = 0.95$. Since online simulation is time-consuming, we use 64 CPU cores and 4 L40S GPUs, requiring 16 hours in total. During early epochs, the safety rate rises sharply while the overtake rate drops, reflecting conservative driving. As training continues, the overtake rate steadily increases while maintaining high safety, demonstrating the progressive acquisition of driving skills (Fig.~\ref{fig:training-curves}(b)).

\subsection{Single-Vehicle Timed Trials}
\label{sec:single-vehicle-timed-trials}

We first conduct single-vehicle timed trials to benchmark navigation capability, reporting lap times and mean speeds in Table~\ref{tab:single_vehicle}. Among prior methods, Zhang--Loidl completes only one track and collides on the rest. Steiner completes all tracks under its official 2.0~m/s speed limit, but collides at even slightly higher speeds. Dreamer finishes all tracks, but averages only around 3.0~m/s. TinyLidarNet is collision-free and notably faster. Predictive Spliner is the fastest, achieving performance comparable to our map-based expert.

Compared to the expert and Predictive Spliner, which both rely on track knowledge and optimized reference racelines, our map-free BC policy matches their lap times and mean speeds. The PPO policy improves performance even further, achieving mean speeds above 8.5~m/s across all tracks. We attribute these gains to online exploration, through which the policy acquires nuanced driving skills to operate at higher speeds, beyond the control limits of the expert. Additionally, it exhibits behaviors aligned with established principles of motorsport driving~\cite{mitchell2004driving}. One example is corner-exit track-out, in which the vehicle moves from the inside to the outside of a curve to minimize speed loss. Another example is corner-exit acceleration and corner-entry braking, in which the vehicle accelerates immediately upon exiting a curve, maintains high speed along the straight, and brakes sharply just before entering the next turn.

\subsection{Head-to-Head Racing}

We evaluate head-to-head performance using safety and overtake rates (Table~\ref{tab:head_to_head}). Safety denotes the proportion of collision-free episodes (encompassing both overtakes and car-following), while the overtake rate measures the percentage of successful passes. Prior methods restricted to low speeds are excluded because they cannot catch up even to the slowest opponent. Predictive Spliner only follows the opponent without attempting an overtake, as its implementation requires an entire lap of car-following to model opponent behavior before initiating a pass. TinyLidarNet remains the only method to execute overtakes, although its success is moderate and falls short of the map-based expert. These results underscore the difficulty of head-to-head racing: despite structured opponent behaviors, prior methods largely fail to execute safe and timely overtakes.

Our BC policy adopts a more conservative racing strategy than the expert, achieving lower overtake rates. However, the conservatism does not translate into improved safety. Such discrepancy stems from the distribution shift inherent to offline imitation learning: while the network mimics demonstrated controls along expert trajectories, it lacks supervision to recover from state deviations that arise during vehicle interactions. Consequently, the policy remains effective for single-vehicle trials but struggles under more complex conditions. This limitation is resolved during the PPO stage, which substantially outperforms the expert in both safety and overtake rates. These results demonstrate that PPO not only bridges the performance gap, but also overcomes the safety flaws of the expert design.

\begin{table}[!t]
\caption{Performance Against Reactive Opponents on Novel Tracks}
\label{tab:ftg_unseen}
\centering
\footnotesize
\setlength{\tabcolsep}{3.5pt}
\begin{tabular*}{\columnwidth}{@{\extracolsep{\fill}}l|ccc@{}}
\hline
\textbf{Method} & \textbf{Safety (\%)} & \textbf{Overtake (\%)} & \textbf{Speed (m/s)} \\
\hline
End2Race (Expert) & \textbf{93.9} & 59.6 & 6.6 \\
End2Race (BC)     & 91.0          & 49.0 & 6.5 \\
\textbf{End2Race (PPO)} & 91.7    & \textbf{85.7} & \textbf{8.3} \\
\hline
\end{tabular*}
\end{table}

\subsection{Generalization to Unseen Opponents}
\label{sec:reactive_opponents}

We evaluate End2Race against an unseen reactive opponent running the Follow-the-Gap (FTG) algorithm. The FTG controller was calibrated to complete all four tracks at high speeds. During scenario execution, it operates at three speed scales, $\alpha_{\mathrm{opp}} \in \{0.5, 0.7, 0.9\}$, capped at 7.5~m/s. To facilitate sustained pursuit before an overtake, we extend the initial separation to 10~m and the scenario duration to 12~s.

Table~\ref{tab:ftg_unseen} reports aggregate performance across the three novel tracks. All three methods maintain safety rates above 90\%, with the map-based expert reaching the highest at 93.9\%. However, overtaking rates for the expert and BC drop because they cannot catch up to the opponent at the 0.9 speed scale. Meanwhile, PPO continues to achieve high overtake rates. Notably, performing overtaking maneuvers at such high speeds is exceptionally challenging due to reduced reaction times and tight stability margins, yet PPO executes them while sustaining high safety. These results indicate that our learned policy generalizes to both novel tracks and unseen opponent behaviors under tight dynamic constraints.

\subsection{Scenario Illustration}

Figure~\ref{fig:scenario-examples} illustrates four cases that highlight the model's behavior during overtaking encounters. All examples are rollouts generated by the PPO policy on the Austin track. We focus primarily on the most challenging configuration ($r_{\mathrm{opp}}=\text{center}$ and $\alpha_{\mathrm{opp}}=0.8$), in which lateral clearance and the speed difference between vehicles are minimized.

\textbf{Overtaking:} Fig.~\ref{fig:scenario-examples}(a) illustrates an overtaking sequence along a straight segment. Upon observing a slower opponent ahead, the ego vehicle initiates a lateral shift while sustaining cruising speed. It maintains a stable lateral clearance alongside the opponent, then smoothly returns to the track center without an abrupt cut-back.

Fig.~\ref{fig:scenario-examples}(b) depicts a cornering overtake. The ego vehicle initially follows the opponent, preparing to pass. When an unexpected sharp curve emerges, it applies moderate braking to sustain higher speed through the turn, then accelerates early at corner exit to complete the pass.

\textbf{Car-Following:} Fig.~\ref{fig:scenario-examples}(c) shows a car-following case. During an overtaking maneuver, the ego vehicle is blocked at the inner corridor by the opponent, forcing it to brake heavily and almost come to a complete stop. Once the opponent proceeds, the ego vehicle accelerates to resume driving.

\textbf{Collision:} Fig.~\ref{fig:scenario-examples}(d) captures a failure case. As the vehicles travel side by side, the track suddenly narrows. With no room along the boundary, the ego vehicle steers toward the opponent, resulting in a collision.

Together, these cases show the tactical racing and overtaking capabilities of the learned policy. Nevertheless, as the policy operates without a map, it cannot predict forward track geometry, which can lead to unfavorable conditions under sudden track constrictions. To address this, future work could explore online adaptation: as the model drives practice laps, it updates its internal neural memory module or adapter weights to encode track layout, and freezes them during the race for more informed decisions~\cite{wijmans2023emergence}.

\subsection{Ablation Study}
\label{sec:ablation}

We conduct an ablation study to examine the effect of design choices in our model architecture, covering two key components: LiDAR normalization and temporal sequence modeling (Table~\ref{tab:ablation}). All variants are trained using behavioral cloning and evaluated on the Austin track.

\subsubsection{LiDAR Normalization}

We compare the proposed sigmoid normalization with two alternatives: no normalization and linear normalization, all sharing the same GRU architecture. The unnormalized model fails to complete a single-vehicle lap. Linear normalization provides a viable policy, but leaves a large performance gap in head-to-head racing. This shows that the sigmoid formulation is crucial in our design for effective driving.

\subsubsection{Temporal Information}

To assess the importance of temporal information, we evaluate an MLP that operates only on the latest observation. The MLP replaces the GRU with a \(210 \rightarrow 420\) linear layer, followed by the action head. The resulting policy exhibits heuristic driving behavior, achieving a higher vehicle speed and overtake rate at the cost of a large drop in safety. These results confirm that temporal cues are essential for modeling vehicle interactions.

\begin{table}[!t]
\vspace{0.15cm}
\caption{Ablation Study on Model Design Choices}
\label{tab:ablation}
\centering
\footnotesize
\setlength{\tabcolsep}{2.5pt}
\begin{tabular*}{\columnwidth}{@{\extracolsep{\fill}}l|cc|cc@{}}
\hline
\multirow{2}{*}{\textbf{Variant}} & \multicolumn{2}{c|}{\textbf{Single-Vehicle}} & \multicolumn{2}{c}{\textbf{Head-to-Head}} \\
& Speed (m/s) & Lap Time (s) & Overtake (\%) & Safety (\%) \\
\hline
No Norm.      & --           & --           & 16.3          & 21.1          \\
Linear Norm.  & 6.4          & 64.9         & 47.0          & 73.3          \\
MLP           & \textbf{6.9} & \textbf{60.9}& 67.6          & 70.4          \\
Transformer   & 6.6          & 63.8         & \textbf{73.1} & 80.6          \\
GRU (BC)      & 6.7          & 62.7         & 62.8          & \textbf{82.8} \\
\hline
\end{tabular*}
\end{table}

\subsubsection{Temporal Modeling}

We evaluate a Transformer~\cite{vaswani2017attention} as an alternative temporal model. The model operates on a 20-token history, corresponding to a span of 0.5~s. Each token represents one timestep and has a dimension of 210. The tokens are processed through two encoder layers with bidirectional self-attention, three attention heads, and a feedforward dimension of 420. The output of the latest token feeds into the action head to produce the control commands. The resulting policy achieves a higher overtake rate, but with a lower safety rate. Additionally, the Transformer incurs four times the inference latency of the GRU. To balance overtaking, safety, and real-time responsiveness, we retain the GRU as our default architecture. Meanwhile, we recognize the Transformer as a practical alternative.

\section{Conclusion}

In this work, we introduce End2Race, an end-to-end learning framework for head-to-head autonomous racing. Our proposed policy achieves low inference latency and delivers competitive racing performance. Future work will deploy the policy in F1Tenth competitions against rival teams.

\bibliographystyle{ieeetr}
\bibliography{F1Tenth}

\end{document}